\documentclass{article}

\usepackage{arxiv}

\usepackage[utf8]{inputenc} % allow utf-8 input
\usepackage[T1]{fontenc}    % use 8-bit T1 fonts
\usepackage{hyperref}       % hyperlinks
\usepackage{url}            % simple URL typesetting
\usepackage{booktabs}       % professional-quality tables
\usepackage{amsfonts}       % blackboard math symbols
\usepackage{nicefrac}       % compact symbols for 1/2, etc.
\usepackage{microtype}      % microtypography
\usepackage{graphicx}
\usepackage{amsmath}
\usepackage{multirow}
\usepackage{listings}
\usepackage{float}
\usepackage{xcolor}

\title{Benchmarking PyCaret AutoML Against IndoBERT Fine-Tuning for
Sentiment Analysis on Indonesian IKN Twitter Data}

\renewcommand{\shorttitle}{PyCaret and IndoBERT for Indonesian IKN Sentiment Analysis}
\renewcommand{\headeright}{}
\renewcommand{\undertitle}{}

\author{
  Mutia Alfi Mayzaroh \\
  Institut Teknologi Sumatera \\
  Lampung Selatan, Indonesia \\
  \texttt{mutia.123410001@student.itera.ac.id} \\
  \And
  Dwi Fitria Ningsih \\
  Institut Teknologi Sumatera \\
  Lampung Selatan, Indonesia \\
  \texttt{dwi.123410005@student.itera.ac.id} \\
  \And
  Nindi Destriani \\
  Institut Teknologi Sumatera \\
  Lampung Selatan, Indonesia \\
  \texttt{nindi.123410006@student.itera.ac.id} \\
  \And
  Martin C.T. Manullang \\
  Institut Teknologi Sumatera \\
  Lampung Selatan, Indonesia \\
  \texttt{martin.manullang@if.itera.ac.id} \\
}

\begin{document}
\maketitle

\begin{abstract}
Sentiment analysis on issues surrounding the development of Indonesia's
new capital city, Ibu Kota Nusantara (IKN), on social media plays an
important role in understanding public opinion in Indonesia. This study
presents a benchmark comparison between a Machine Learning (ML) approach
based on AutoML with PyCaret and a Deep Learning (DL) approach based on
IndoBERT fine-tuning for binary sentiment classification (positive/negative)
on an Indonesian-language Twitter comment dataset related to IKN. The
dataset consists of 1,472 samples (780 negative and 692 positive). In the
ML approach, three algorithms were evaluated through 10-fold
cross-validation: Logistic Regression (LR), Naive Bayes (NB), and Support
Vector Machine (SVM). Logistic Regression achieved the best performance
among the ML models with an average accuracy of 77.57\% and an F1-score of
77.17\%. Meanwhile, in the DL approach, fine-tuning the
\textit{indobenchmark/indobert-base-p1} model using PyTorch for 5 epochs
yielded a test set accuracy of 89.59\% and an F1-score of 89.37\%. The
results show that IndoBERT significantly outperforms all ML models by a
margin of more than 12 percentage points in accuracy, confirming the
superiority of Transformer-based contextual representations in handling
informal Indonesian-language text.
\end{abstract}

\keywords{Sentiment Analysis \and PyCaret \and IndoBERT \and AutoML \and
Indonesian NLP \and IKN \and Transformer}

% -----------------------------------------------
\section{Introduction}
% -----------------------------------------------

Ibu Kota Nusantara (IKN) is Indonesia's capital relocation project, which
has become one of the most widely discussed topics on social media since its
announcement. Various segments of society have expressed their views through
platforms such as Twitter/X, generating a large and diverse volume of
comments. Understanding the distribution of public sentiment toward this
project holds strategic value for policymakers and researchers alike.

In the field of Natural Language Processing (NLP), sentiment analysis has
evolved rapidly from rule-based approaches to machine learning, and now to
\textit{deep learning} \cite{medhat2014sentiment}. The availability of
Indonesian pre-trained language models such as IndoBERT
\cite{koto2020indolem} opens opportunities to apply cutting-edge techniques
to Indonesian, a language that is considered \textit{low-resource}.

This study contributes: (1) an Indonesian-language IKN sentiment dataset of
1,472 manually labeled samples; (2) a comprehensive benchmark of three ML
algorithms (LR, NB, SVM) via PyCaret AutoML; (3) an IndoBERT fine-tuning
implementation using PyTorch; and (4) a comparative analysis answering the
question: \textit{does a Transformer-based DL approach outperform classical
ML for sentiment analysis of informal Indonesian text?}

% -----------------------------------------------
\section{Related Work}
% -----------------------------------------------

Sentiment analysis on Indonesian-language social media text has been widely
studied using various approaches. Alfina et al.~\cite{alfina2017hate} applied
SVM and Naive Bayes for hate speech detection in Indonesian, finding that SVM
outperformed using n-gram features.

Riadi et al.~\cite{riadi2019sentiment} used Naive Bayes and SVM on an
Indonesian Twitter dataset for political sentiment analysis, with SVM
achieving 83\% accuracy using TF-IDF features.

In the context of AutoML, PyCaret provides an integrated framework that
simplifies ML experimentation through automated preprocessing, model
comparison, and hyperparameter tuning~\cite{ali2020pycaret}, and has become
increasingly popular for text classification tasks~\cite{ramadhani2023automl}.

In deep learning, the Transformer architecture introduced by Vaswani
et al.~\cite{vaswani2017attention} revolutionized NLP through the
\textit{self-attention} mechanism. BERT~\cite{devlin2019bert} became the
foundation for domain-specific pre-trained models, including
IndoBERT~\cite{koto2020indolem}, which was trained on a large Indonesian
corpus and has been shown to outperform ML baselines on Indonesian NLP
tasks~\cite{wilie2020indonlu}.

A comparative study of ML vs.\ DL on Indonesian text by Salsabila
et al.~\cite{salsabila2022comparison} showed that BERT-based models
consistently outperform traditional ML on e-commerce sentiment (F1
improvement of 10--15 points). Similar findings were reported by Kurniawan
and Louvan~\cite{kurniawan2023bert} on news sentiment (IndoBERT F1-macro
89.3\% vs.\ SVM 76.8\%).

Research on public understanding of government policy through social media is
growing~\cite{ibrahim2023social}, with the need for accurate Indonesian
sentiment analysis becoming increasingly urgent amid the surge in social media
content~\cite{purwarianti2023indonesian}.

% -----------------------------------------------
\section{Dataset}
% -----------------------------------------------

\subsection{Dataset Description}

The dataset used in this study was sourced from Indonesian-language Twitter/X
comments on the topic of Ibu Kota Nusantara (IKN). Data were collected
through keyword-based searches related to IKN and manually labeled into two
classes: \textbf{positive} and \textbf{negative}.

\subsection{Dataset Statistics}

The final dataset after preprocessing consists of 1,472 samples with the
following class distribution shown in Table~\ref{tab:dataset}.

\begin{table}[H]
\centering
\caption{Class Distribution of the IKN Sentiment Dataset}
\label{tab:dataset}
\begin{tabular}{lcc}
\toprule
\textbf{Class} & \textbf{Number of Samples} & \textbf{Percentage} \\
\midrule
Negative & 780 & 53.0\% \\
Positive & 692 & 47.0\% \\
\midrule
\textbf{Total} & \textbf{1,472} & \textbf{100\%} \\
\bottomrule
\end{tabular}
\end{table}

Text length statistics (in characters) show a mean of 50.3 characters
(std=38.5), a minimum of 2 characters, a median of 39 characters, and a
maximum of 205 characters. This reflects the characteristic brevity and
informality of social media comments.

\subsection{Preprocessing}

The preprocessing steps include:
\begin{enumerate}
    \item \textbf{Lowercasing}: All text is converted to lowercase.
    \item \textbf{Removal of missing values}: Rows with \texttt{NaN} values
    are dropped.
    \item \textbf{Text length computation}: An additional
    \texttt{text\_length} feature is calculated.
\end{enumerate}

The cleaned data is saved as \texttt{clean\_data.csv} for use in the
subsequent modeling stage.

% -----------------------------------------------
\section{Methodology}
% -----------------------------------------------

\subsection{Machine Learning Pipeline with PyCaret}

PyCaret version 3.x was used as the AutoML framework for machine learning
experiments. The \texttt{setup()} function was configured with the
\texttt{text} column as the text feature and \texttt{label} as the target,
with \texttt{session\_id=42} for reproducibility. PyCaret automatically
performs feature extraction using TF-IDF from the text column.

Three ML algorithms were compared:
\begin{enumerate}
    \item \textbf{Logistic Regression (LR)}: A linear model with L2
    regularization, \texttt{C=1.0}, \texttt{lbfgs} solver, and
    \texttt{max\_iter=1000}.
    \item \textbf{Naive Bayes (NB)}: A probabilistic classifier based on
    Bayes' theorem, well-suited for high-frequency text data.
    \item \textbf{Support Vector Machine (SVM)}: SVM with a linear kernel for
    high-dimensional text classification.
\end{enumerate}

Evaluation was performed using 10-fold stratified cross-validation. The best
model was then tuned using \texttt{tune\_model()} with Optuna-based
hyperparameter search (10 candidates, 10 folds), and finalized with
\texttt{finalize\_model()}.

\subsection{Deep Learning Pipeline: IndoBERT Fine-Tuning}

For the DL approach, the \textit{indobenchmark/indobert-base-p1}
model~\cite{koto2020indolem} was fine-tuned using PyTorch. IndoBERT is a
BERT-based model with 12 Transformer layers, 768 hidden dimensions, 12
attention heads, and approximately 111 million parameters. Classification was
performed by adding a linear layer (\textit{classifier head}) on top of the
\texttt{[CLS]} token representation.

\textbf{Hyperparameter configuration:}
\begin{itemize}
    \item \texttt{MAX\_LEN} = 128 tokens
    \item \texttt{BATCH\_SIZE} = 16
    \item \texttt{EPOCHS} = 5
    \item \texttt{Learning Rate} = $2 \times 10^{-5}$
    (AdamW, weight\_decay=0.01)
    \item Scheduler: linear warmup (10\% of total steps), followed by
    linear decay
    \item Gradient clipping: max\_norm = 1.0
\end{itemize}

\textbf{Data split:}
\begin{itemize}
    \item Train: 1,030 samples (70\%)
    \item Validation: 221 samples (15\%)
    \item Test: 221 samples (15\%)
\end{itemize}

A stratified split was used to preserve the class distribution in each
partition. The best model (based on highest val\_acc) was saved during
training.

\subsection{AI Prompts Used}

In developing this pipeline, an AI assistant (Claude) was used to assist
with code writing. The following are the main prompts used:

\begin{lstlisting}[basicstyle=\small\ttfamily, breaklines=true, frame=single,
                   caption={Prompt for EDA and Preprocessing}]
"Create a Python notebook for EDA and preprocessing of a sentiment
CSV dataset with 'text' and 'label' columns. Include: load data,
check label distribution, compute text length statistics, lowercase
text, drop NaN values, and save clean_data.csv."
\end{lstlisting}

\begin{lstlisting}[basicstyle=\small\ttfamily, breaklines=true, frame=single,
                   caption={Prompt for PyCaret AutoML}]
"Create a PyCaret notebook for binary text sentiment classification.
Use pycaret.classification with text_features=['text']. Compare
Logistic Regression, Naive Bayes, and SVM models using create_model
and compare_models. Run tune_model on the best model and
finalize_model, then save with save_model."
\end{lstlisting}

\begin{lstlisting}[basicstyle=\small\ttfamily, breaklines=true, frame=single,
                   caption={Prompt for IndoBERT Fine-Tuning}]
"Create a PyTorch notebook for fine-tuning IndoBERT
(indobenchmark/indobert-base-p1) on a binary Indonesian sentiment
classification task. Include: custom Dataset class, DataLoader,
training loop with AdamW optimizer, linear warmup scheduler,
gradient clipping, per-epoch evaluation (loss, accuracy), save best
model based on val_accuracy, evaluate test set with
classification_report and confusion matrix, and an inference
function for new text."
\end{lstlisting}

% -----------------------------------------------
\section{Experiments}
% -----------------------------------------------

\subsection{Experimental Setup}

The ML experiments (PyCaret) were run on an Intel Core i5 CPU using Python
3.9. The DL experiments (IndoBERT) were run on the same CPU without GPU
acceleration, using PyTorch 2.x and the Hugging Face Transformers library.

\subsection{Evaluation Metrics}

The following metrics were used:
\begin{itemize}
    \item \textbf{Accuracy}: The proportion of correct predictions out of all
    samples.
    \item \textbf{Precision}: The correctness of positive predictions.
    \item \textbf{Recall}: The completeness of positive class detection.
    \item \textbf{F1-Score}: The harmonic mean of precision and recall.
    \item \textbf{AUC-ROC}: The area under the ROC curve (ML models only).
\end{itemize}

For ML models, all metrics were averaged across 10-fold cross-validation. For
IndoBERT, evaluation was performed on the test set that was never seen during
training or validation.

% -----------------------------------------------
\section{Results and Discussion}
% -----------------------------------------------

\subsection{Machine Learning Model Results (PyCaret)}

Table~\ref{tab:ml_results} summarizes the average 10-fold cross-validation
performance for the three ML models compared.

\begin{table}[H]
\centering
\caption{ML Model Performance Comparison via PyCaret (10-Fold CV)}
\label{tab:ml_results}
\begin{tabular}{lccccc}
\toprule
\textbf{Model} & \textbf{Accuracy} & \textbf{AUC} & \textbf{Precision}
& \textbf{Recall} & \textbf{F1} \\
\midrule
Logistic Regression & 0.7757 & 0.8563 & 0.7848 & 0.7757 & 0.7717 \\
Naive Bayes         & 0.7058 & 0.7110 & 0.7124 & 0.7058 & 0.7053 \\
SVM (Linear)        & 0.5272 & 0.6427 & 0.5739 & 0.5272 & 0.4673 \\
\bottomrule
\end{tabular}
\end{table}

Logistic Regression proved to be the best ML model with an accuracy of
77.57\% and F1 of 77.17\%. This performance is higher than Naive Bayes
(70.58\%) and significantly outperforms Linear SVM (52.72\%). The low SVM
performance is likely due to the high lexical variation in informal social
media text, making PyCaret's default TF-IDF features suboptimal for a linear
kernel SVM without extensive tuning. After the \texttt{tune\_model} process
on SVM, performance improved to 71.17\%.

\subsection{Deep Learning Model Results (IndoBERT)}

Table~\ref{tab:dl_results} presents the IndoBERT evaluation metrics on the
test set.

\begin{table}[H]
\centering
\caption{IndoBERT Fine-Tuning Performance on the Test Set}
\label{tab:dl_results}
\begin{tabular}{lcccc}
\toprule
\textbf{Class} & \textbf{Precision} & \textbf{Recall} & \textbf{F1}
& \textbf{Support} \\
\midrule
Negative & 0.86 & 0.96 & 0.91 & 117 \\
Positive & 0.95 & 0.83 & 0.88 & 104 \\
\midrule
\textbf{Accuracy} & \multicolumn{4}{c}{\textbf{0.8959}} \\
\textbf{Macro Avg} & 0.90 & 0.89 & 0.89 & 221 \\
\textbf{Weighted Avg} & 0.90 & 0.90 & 0.90 & 221 \\
\bottomrule
\end{tabular}
\end{table}

IndoBERT achieved a test set accuracy of 89.59\% with an F1-macro of 89.37\%.
During training, the best model was obtained at the first epoch with a
val\_accuracy of 85.07\%, indicating overfitting in subsequent epochs as
evidenced by increasing val\_loss. This phenomenon is common when fine-tuning
BERT on medium-sized datasets.

\begin{figure}[H]
\centering
\includegraphics[width=\linewidth]{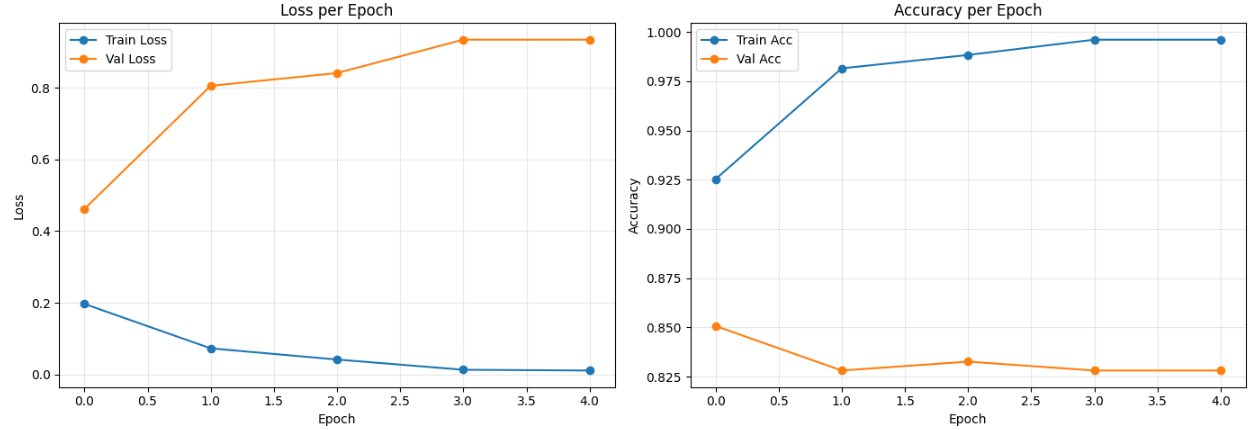}
\caption{Loss and Accuracy Curves per Epoch --- IndoBERT Fine-Tuning}
\label{fig:training}
\end{figure}

Figure~\ref{fig:training} shows the training and validation curves of
IndoBERT over 5 epochs. The train loss continuously decreases toward zero,
while the val loss increases after the first epoch, indicating overfitting.
Conversely, train accuracy reaches nearly 100\%, while val accuracy
stabilizes in the range of 82--85\%.

\begin{figure}[H]
\centering
\includegraphics[width=0.6\linewidth]{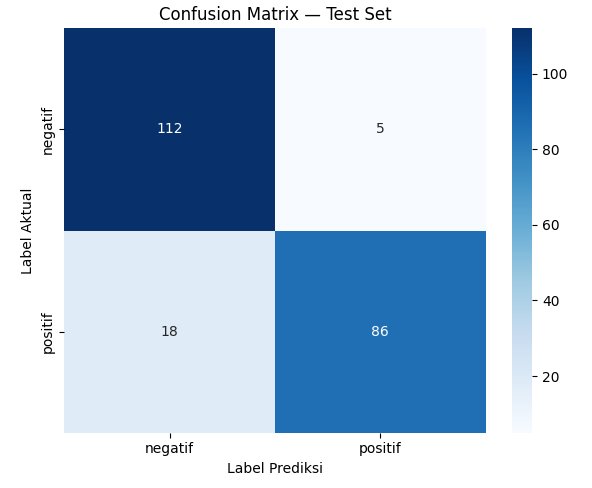}
\caption{Confusion Matrix of IndoBERT on the Test Set}
\label{fig:cm}
\end{figure}

Figure~\ref{fig:cm} shows the confusion matrix on the test set. The model
successfully classified 112 out of 117 negative samples (recall 95.7\%) and
86 out of 104 positive samples (recall 82.7\%). Misclassifications were more
frequent in the positive class being predicted as negative (18 samples),
likely due to positive expressions using indirect language or irony.

\subsection{Comprehensive Benchmark}

Table~\ref{tab:benchmark} summarizes the comparison of all models side by
side.

\begin{table}[H]
\centering
\caption{Full Benchmark: ML vs.\ DL}
\label{tab:benchmark}
\begin{tabular}{llccc}
\toprule
\textbf{Approach} & \textbf{Model} & \textbf{Accuracy} & \textbf{F1 (Macro)}
& \textbf{Note} \\
\midrule
\multirow{3}{*}{ML (PyCaret)}
  & Logistic Regression & 0.7757 & 0.7717 & Best ML model \\
  & Naive Bayes         & 0.7058 & 0.7053 & \\
  & SVM (Tuned)         & 0.7117 & 0.6945 & After tune\_model \\
\midrule
DL (PyTorch)
  & IndoBERT (fine-tuned) & \textbf{0.8959} & \textbf{0.8937}
  & \textbf{Best overall} \\
\bottomrule
\end{tabular}
\end{table}

\subsection{Comparative Analysis}

IndoBERT consistently outperforms all ML models with an accuracy margin of
\textbf{+12.02 percentage points} compared to the best ML model (LR). This
superiority can be explained through several factors:

\begin{enumerate}
    \item \textbf{Contextual representation}: IndoBERT understands word
    context bidirectionally, whereas TF-IDF only counts word frequencies
    without context.
    \item \textbf{Pre-training on Indonesian}: The linguistic knowledge
    acquired by IndoBERT from pre-training on a large corpus helps it
    understand slang, abbreviations, and informal writing styles
    characteristic of Indonesian social media.
    \item \textbf{WordPiece tokenization}: IndoBERT's tokenizer handles
    out-of-vocabulary (OOV) words more effectively than bag-of-words
    representations.
\end{enumerate}

However, IndoBERT requires significantly greater computational resources
($\sim$111M parameters vs.\ lightweight ML models) and considerably longer
training time, especially without a GPU. ML models, particularly Logistic
Regression, remain a pragmatic choice when computational resources are
limited.

% -----------------------------------------------
\section{Conclusion}
% -----------------------------------------------

This paper presents a benchmark between Machine Learning (PyCaret AutoML) and
Deep Learning (IndoBERT fine-tuning) for sentiment analysis of
Indonesian-language Twitter comments about IKN. The following conclusions are
drawn from the experiments:

\begin{enumerate}
    \item \textbf{IndoBERT is the best overall model} with an accuracy of
    89.59\% and F1-macro of 89.37\% on the test set, outperforming all ML
    models.
    \item \textbf{Logistic Regression is the best ML model} with an accuracy
    of 77.57\% via 10-fold CV, outperforming Naive Bayes and SVM in the
    PyCaret AutoML context.
    \item \textbf{Transformer-based Deep Learning} significantly outperforms
    classical ML for informal Indonesian text, with an accuracy improvement
    of more than 12 percentage points over the best ML model.
    \item \textbf{Trade-offs must be considered}: IndoBERT requires far
    greater computational resources, so model selection should be adapted to
    available resources and tolerance for inference latency.
\end{enumerate}

For future work, it is recommended to explore data augmentation techniques to
address potential overfitting in IndoBERT, as well as to compare with lighter
models such as DistilBERT or LSTM-based architectures.

% -----------------------------------------------

\end{document}